\documentclass[runningheads]{llncs}

\usepackage[mobile]{eccv}
\pdfoutput=1

\usepackage{eccvabbrv}
\usepackage{amssymb}
\usepackage{graphicx}
\usepackage{booktabs}
\usepackage{tabularx}
\usepackage{multirow}
\usepackage{dsfont}
\usepackage{pifont}
\usepackage[misc]{ifsym}

\usepackage[accsupp]{axessibility}  
\newcommand{\cmark}{\ding{51}}

\newcommand{\thename}{\textsc{MIM4D}}

\usepackage[pagebackref,breaklinks,colorlinks,citecolor=eccvblue]{hyperref}

\usepackage{orcidlink}

\begin{document}

\title{\thename{}: Masked Modeling \\with Multi-View Video for Autonomous Driving Representation Learning} 

\titlerunning{MIM4D}

\author{Jialv Zou\inst{1,\star,\diamond}\and
Bencheng Liao\inst{1,2,\star,\diamond } \and
Qian Zhang\inst{3} \and \\
Wenyu Liu\inst{1} \and
Xinggang Wang\inst{1,\textrm{\Letter}}
}


\authorrunning{J. Zou, X. Wang et al.}

\institute{School of EIC, Huazhong University of Science \& Technology \quad \and
Institute of Artificial Intelligence, Huazhong University of Science \& Technology \and
Horizon Robotics\\
}

\maketitle
\let\thefootnote\relax\footnotetext{$^\star$ Equal contribution; $^\diamond$ Interns of Horizon Robotics when doing this work; $^\textrm{\Letter}$ Corresponding author: \texttt{xgwang@hust.edu.cn}}

\begin{abstract}
Learning robust and scalable visual representations from massive multi-view video data remains a challenge in computer vision and autonomous driving. Existing pre-training methods either rely on expensive supervised learning with 3D annotations, limiting the scalability, or focus on single-frame or monocular inputs, neglecting the temporal information. We propose \thename{}, a novel pre-training paradigm based on dual masked image modeling (MIM). \thename{} leverages both spatial and temporal relations by training on masked multi-view video inputs. It constructs pseudo-3D features using continuous scene flow and projects them onto 2D plane for supervision. To address the lack of dense 3D supervision, \thename{} reconstruct pixels by employing 3D volumetric differentiable rendering to learn geometric representations. We demonstrate that \thename{} achieves state-of-the-art performance on the nuScenes dataset for visual representation learning in autonomous driving. It significantly improves existing methods on multiple downstream tasks, including BEV segmentation ($8.7\%$ IoU), 3D object detection ($3.5\%$ mAP), and HD map construction ($1.4\%$ mAP). Our work offers a new choice for learning representation at scale in autonomous driving. Code and models are released at \url{https://github.com/hustvl/MIM4D}.

\end{abstract}

\section{Introduction}
\label{sec:intro}

There is a growing interest in pursuing vision-centric autonomous driving. A bunch of works focus on extracting Bird’s-Eye-View (BEV) features from multi-view input images to address various downstream tasks in a supervised setting. However, compared to the limited annotated data used in supervised learning, there is a huge amount of unlabeled data.  Exploring how to exploit and perform representation learning from the unlabeled data to benefit various downstream tasks in autonomous driving is challenging and still under-explored.

Existing pre-training methods for 3D driving scenes can be broadly categorized into depth-supervised methods and  NeRF-based methods. The depth estimation based method~\cite{park2021pseudo} using the geometric relationship between Image-LiDAR pairs to obtain pixel-wise depth maps for the images and utilizes them as supervisory signals to pre-train the model. However, this method only extracts depth from a single view, neglecting the geometric information from multiple views.
Inspired by the success of NeRF \cite{mildenhall2021nerf} in novel view synthesis, it has been adopted as a novel self-supervised learning method. Specifically, these methods construct a NeRF-style 3D volume representation from multi-view images and utilize advanced volume rendering techniques to generate 2D renderings, then use 2d labels as supervisions. For example, OccNeRF \cite{zhang2023occnerf} utilizes a conventional occupancy network to obtain a 3D representation of voxels from the multi-view images. Then, it renders the 3D representation back to the multi-view images using NeRF. with 2D segmentation labels provided by a frozen SAM \cite{kirillov2023segment} model. While this method avoids the cumbersome task of annotating 3D labels, it still falls into the supervised paradigm, and the performance is limited by the quality of the annotations provided by the SAM model. UniPAD \cite{yang2023unipad} combines MAE \cite{he2022masked} with a neural rendering network, it takes the masked images as input and aims to reconstruct the missing pixels on the rendered 2D image via neural rendering.

However, the aforementioned methods lack temporal modeling, which plays a crucial role in end-to-end autonomous driving systems, \eg, UniAD \cite{hu2023planning} and VAD series \cite{jiang2023vad, chen2024vadv2}. Perception is the upstream component of an autonomous driving system, aiming to serve prediction and planning, which are the ultimate goals.
Therefore, precise perception of the dynamic scene flow and moving objects is of utmost importance. Due to the lack of temporal modeling, such methods are insufficient for pre-training the end-to-end system.

\begin{figure}[t!]
  \centering
  \includegraphics[width=12cm,keepaspectratio]{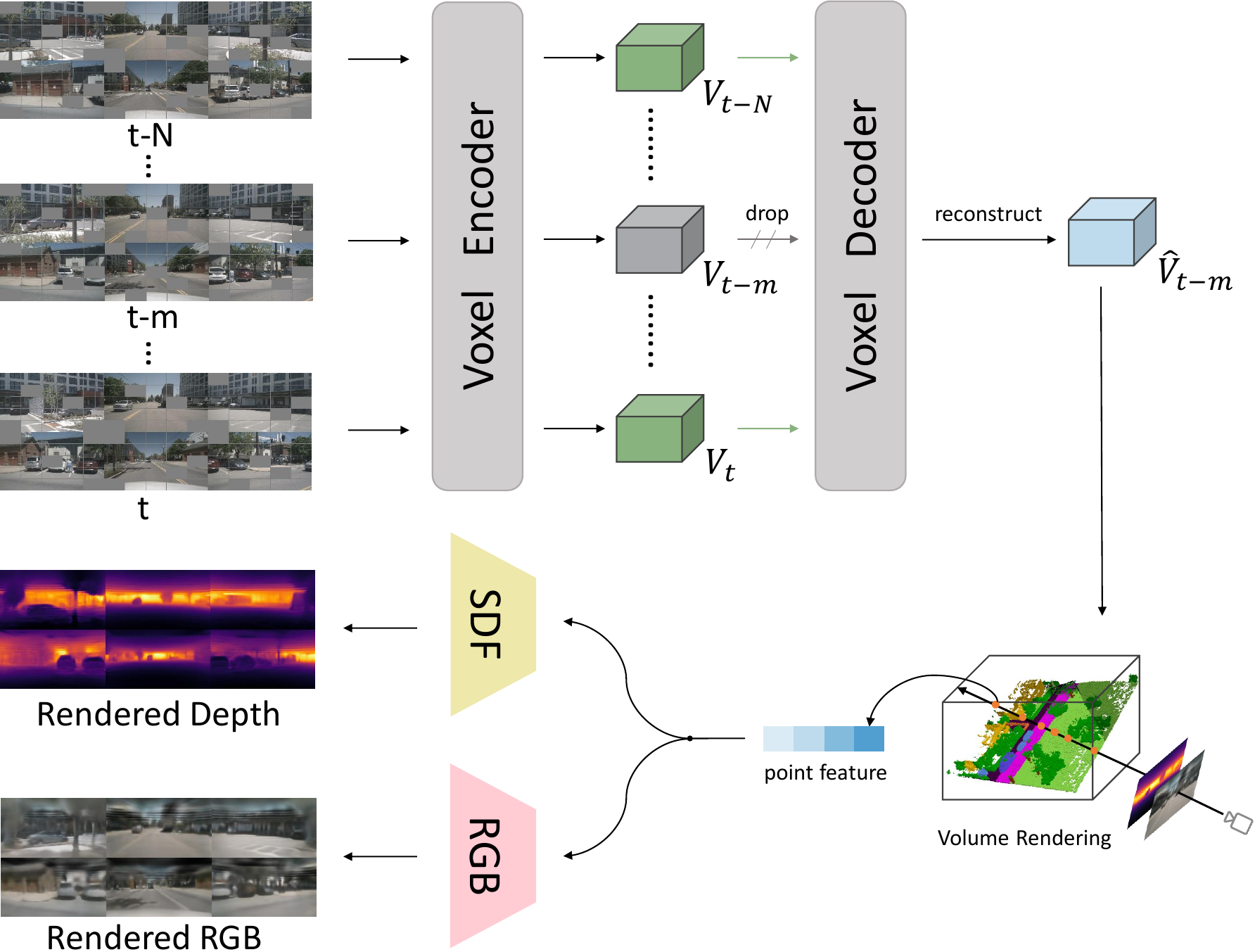}
  \caption{The overall architecture of \thename{}. We take the masked multi-frame multi-view images as input and utilize a voxel encoder to obtain the voxel feature sequence. Specifically, we utilize an image backbone network to extract hierarchical features, following which a view transformation module is employed to convert the multi-view features into representations in 3D space. After that, we drop the voxel feature of the m-th frame and reconstruct it base on the remaining voxel feature sequence using the voxel decoder module. Finally, the reconstructed $\hat{\textbf{V}}_{t-m}$ is fed into the volume-based neural rendering decoder, where it is projected back onto a 2D plane. The 2D pixel-wise RGB and depth are used as supervision.}
  \label{The overall architecture}
\end{figure}

Motivated by the success of MAE \cite{he2022masked} in 2D image tasks and the limitations of current pre-training methods in 3D space, we aspire to extend MAE to 3D visual tasks. In this paper, we propose a novel approach \thename{}, which is a dual masked image modeling (MIM) architecture in both temporal and spatial domains, it synergizes temporal modeling with volume rendering as a new pre-training paradigm. The architecture of our method is shown in Fig.~\ref{The overall architecture}. In the spatial domain, using the masked multi-view video as input, we employ 3D differentiable volume rendering to reconstruct the missing geometric information. In the temporal domain, we randomly drop the features of one frame and reconstruct it from the remaining feature sequence. By incorporating temporal modeling, we aim to address the limitations of volume rendering pre-training methods in capturing dynamic scene flow, making them mutually complementary and synergistic.

We conduct extensive experiments  to evaluate the effectiveness of our pre-training method \thename{} on the nuScenes dataset \cite{caesar2020nuscenes}. Firstly, we compare our method with previous visual representation learning methods. Experimental results show that our approach outperforms both supervised and unsupervised representation learning methods. Upon leveraging our pre-trained model as the backbone, we observe a remarkable increase of 9.2 mAP and 6.6 NDS compared to the baseline, which is pre-trained on ImageNet \cite{deng2009imagenet}. Furthermore, our model achieves an improvement over the previous state-of-the-art representation learning method UniPAD \cite{yang2023unipad}, with a gain of 0.8 mAP and a boost of 1.1 NDS. These results provide evidence that extending MAE to the temporal dimension as a paradigm for representation learning is effective. Then, we evaluate our pre-trained model on a wide range of downstream tasks to ensure the generality of our approach. Experimental results demonstrate that the backbone pre-training with our method, exhibits significant improvements in various downstream tasks, compared to the baselines. For BEV segmentation task, CVT \cite{zhou2022cross} achieved a notable $8.7\%$ increase in IoU when utilizing our pre-trained model. For the 3D object detection, our method yields $2.6\%$ NDS improvement to PETR \cite{liu2022petr} and $3.5\%$ mAP improvement to BEVDet4D \cite{huang2022bevdet4d}, respectively. Even when applied to the current best-performing detector Sparse4Dv3 \cite{lin2023sparse4d}, which constitutes the state-of-the-art approach with fine-grained temporal modeling and 
applied depth estimation as an additional auxiliary task, our method still demonstrates a 0.6 non-trivial improvement on the NDS metric. For HD map construction task, our method consistently brings an average improvement of over $1.3\%$ mAP to the state-of-the-art method MapTR~\cite{liao2022maptr}.

Our contributions can be summarized as follows:
\begin{itemize}

\item We extend masked image modeling (MIM) to 4D space by leveraging continuous scene flow to construct the dropped voxel feature, thus modeling the temporal information and enhancing the model's capability to capture the motion flow within dynamic scenes.

\item We employ 3D volumetric differentiable rendering to project voxel features onto a 2D plane for supervision, which implicitly provides continuous supervision signals to the model for learning 3D structures, avoiding the need of expensive 3D annotations.

\item We conduct extensive experiments on the nuScenes dataset. Our pre-training method outperforms previous supervised and unsupervised representation learning approaches and performs well across a wide range of downstream tasks and various backbone architectures. This provides convincing evidence for the effectiveness and universality of \thename{}.

\end{itemize}

\section{Related Work}
\paragraph{\textbf{Self-supervised Image Representation Learning.}}
Self-supervised representation learning in 2D images has achieved great success. SimCLR~\cite{simclr} proposes a contrastive learning framework to largely boost the downstream performance of self-supervised learning. MoCo~\cite{moco} builds a dynamic dictionary with a queue and a moving-averaged encoder to facilitate contrastive unsupervised learning. A series of works~\cite{chen2020improved,fan2021multiscale,caron2021emerging,radford2021learning}, follow the contrastive paradigm. Recently, another line of works~\cite{fang2023corrupted, beit, he2022masked,fang2023eva,xie2022simmim,li2023scaling,spark}, such as BEiT~\cite{beit} and MAE~\cite{he2022masked},  learn representations from corrupted images by reconstructing the masked image patches. The following works~\cite{wang2023videomae,videomae,maest}, such as VideoMAE~\cite{videomae} and MAE-ST~\cite{maest}, further advance this paradigm in the temporal domain. However, these MAE-style approaches remain confined to structural learning within 2D images, directly applying them to autonomous driving can not enhance the model's geometric perception capability in 3D space, which is key for downstream perception, motion prediction, and planning tasks. Our proposed MIM4D is aimed at filling this gap.

\paragraph{\textbf{Neural Rendering.}}
NeRF~\cite{nerf} is a pioneering work, utilizing neural networks to differentially render images from 3D scene representation. A lot of works~\cite{dvgo,muller2022instant} follow the paradigm, focusing on fast and accurate reconstruction. mip-NeRF \cite{barron2021mip} enhances NeRF's ability to represent fine details by efficiently rendering anti-aliased conical frustums instead of rays, effectively reducing bothersome aliasing artifacts. NeuS~\cite{neus} proposes to represent a surface as the zero-level set of  a signed distance function (SDF) and achieves high-quality reconstruction with the SDF representation. TiNeuVox~\cite{tineuvox} advances NeRF in 4D reconstruction by representing scenes with time-aware voxel features. Another line of work~\cite{huang2024nerf,xu2023nerf,xu2023mononerd} utilizes NeRF to perform representation learning for perception tasks. NeRF-Det \cite{xu2023nerf} applies an extra NeRF branch to learn a geometry-aware volumetric representation for 3D detection. MonoNeRD \cite{xu2023mononerd} also introduces NeRF representation into the traditional detection pipeline to achieve superior performance. Above works either use NeRF to perform reconstruction, or apply NeRF in supervised detection tasks as auxiliary branch. Different from them, we extend NeRF into the self-supervised representation learning domain, using it as pretext task for scalable pre-training.

\paragraph{\textbf{Representation Learning for Autonomous Driving}} 
Autonomous driving requires accurate geometrical representation. A bunch of works~\cite{chen20224dcontrast,li2022closer,liang2021exploring,liu2022masked,pang2022masked,tian2023geomae} follow either the masked modeling paradigm or the contrastive paradigm to perform representation learning in point cloud modality. Ponder \cite{huang2023ponder} further introduces neural rendering into point cloud pre-training. 
However, there is a lack of effort to perform image-based representation learning for autonomous driving. The most well-known approach is DD3D~\cite{park2021pseudo}, which performs depth pre-training to largely benefit the downstream detection performance. Recently, GeoMIM~\cite{Liu2023GeoMIMTB} applies masked image modeling on BEV to transfer the knowledge from LiDAR model to multi-view camera-based 3D detection. UniPAD~\cite{yang2023unipad} proposes to utilize neural rendering to further enhance the representation learning, OccNeRF \cite{zhang2023occnerf} utilizes frozen SAM \cite{kirillov2023segment} to provide 2D labels as supervision for the neural rendering network. These above representation learning methods either only focus on a single view, neglecting the geometric information derived from multiple views, or lack modeling of temporal information. Our proposed \thename{} fully exploits the massive multi-view video in autonomous driving.

\section{Method}

In this section, we elaborate on our \thename{}. We begin with an overview of our work in \cref{subsec:Overview}, and subsequently delve into Voxel Encoder in \cref{subsec:Encoder},  Voxel Decoder in \cref{subsec:Temporal}, Neural Rendering in \cref{subsec:Decoder} and Self-Supervised Loss in \cref{subsec:Loss}, respectively.

\subsection{Overview}
\label{subsec:Overview}
The framework of \thename{} is shown in Fig.~\ref{The overall architecture}, which contains three parts: \textbf{(a)} a voxel encoder, composed of an image backbone and an image-view transform module, which is used to extract 3D volume features $\textbf{V}$ from masked multi-frame multi-view images $\left \{ \textbf{I}_{t-N}, ...,\textbf{I}_{t} \right \} $; \textbf{(b)} a voxel decoder, given the voxel feature sequence $\left \{ \textbf{V}_{t-N}, ...,\textbf{V}_{t} \right \} $, it randomly drops one voxel $\textbf{V}_{t-m}$ and utilizes the remaining voxel sequence to reconstruct it; \textbf{(c)}
a neural rendering network, which performs volume rendering from voxel features and optimizes the whole network with 2D pixel-wise color $C_{RGB}$ and depth value $D_{depth}$. The entire process can be formulated as:

\begin{equation}
\begin{split}
\left \{ \textbf{V}_{t-N}, ...,\textbf{V}_{t} \right \} &= Voxel \ Encoder(\left \{ \textbf{I}_{t-N}, ...,\textbf{I}_{t} \right \}), \\
\hat{\textbf{V}}_{t-m} = Voxel& \ Decoder(\left \{ \textbf{V}_{t-N}, ...,\textbf{V}_{t} \right \} - \textbf{V}_{t-m}), \\
\hat{C}_{RGB}, \hat{D}_{depth} &= Neural \ Rendering(\hat{\textbf{V}}_{t-m}). \\
\end{split}
\end{equation}

\subsection{Voxel Encoder}
\label{subsec:Encoder}
Given the multi-view images $I$, inspired by MAE\cite{he2022masked}, we randomly drop some patches from these inputs and replace traditional convolutions in image backbone with sparse convolutions \cite{liu2015sparse} as in SparK \cite{tian2023designing}, enabling it to handle masked images, similar to Transformers. By using this approach, we can guide the model to learn the global structure of the image, thereby enhancing the model performance and generalization capability. Subsequently, we employ lift-split-shoot (LSS) \cite{philion2020lift} as the image-view transform module to construct 3D volume $\textbf{V} \in R^{C\times Z\times H\times W} $ from multi-view image features, where C, W, H, Z represent the channel number, the number of x/y/z dimensions in the 3D space respectively.

\subsection{Voxel Decoder}
\label{subsec:Temporal}
\begin{figure}[t!]
  \centering
  \includegraphics[width=12cm,keepaspectratio]{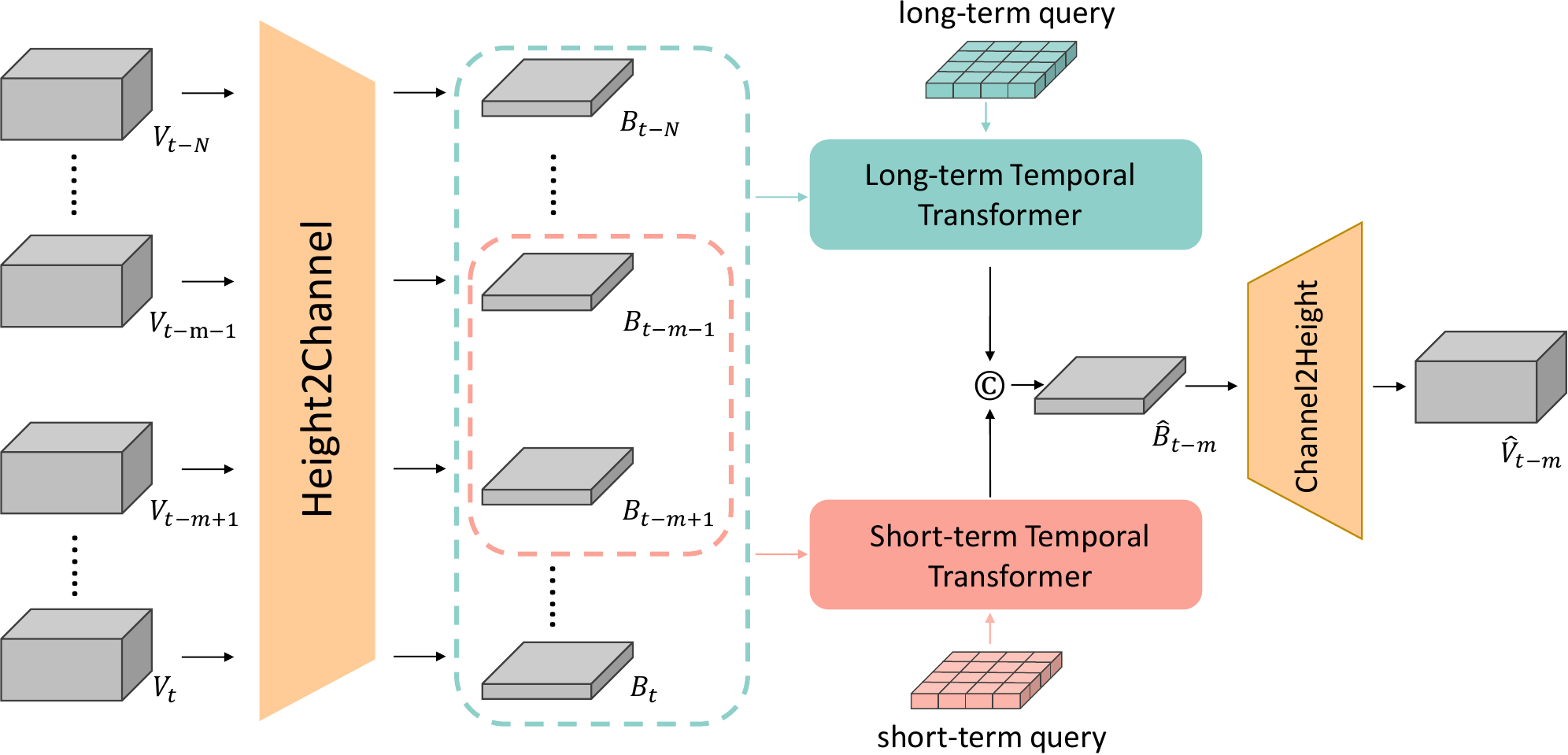}
  \caption{The architecture of voxel decoder. Taking the voxel sequence as input, it is first processed by the Height2Channel module to convert voxel features into BEV features, temporarily compressing the height dimension. Next, we use the long-term temporal transformer and the short-term temporal transformer to extract information from the remaining BEV feature sequence, which are then fused together to reconstruct the dropped BEV feature. The symbol © denotes feature concatenation and fusion. Finally, we employ the Channel2Height module to convert the reconstructed BEV feature back to voxel feature. }
  \label{temporal transformer}
\end{figure}

Through the voxel encoder, we obtain a voxel feature sequence $\left \{ \textbf{V}_{t-N}, ...,\textbf{V}_{t} \right \} $. To model temporal dependencies, we randomly drop one voxel feature $\textbf{V}_{t-m}$ and leverage the remaining voxel features to reconstruct it, similar to HoP \cite{zong2023temporal}. The architecture of voxel decoder is shown in Fig.~\ref{temporal transformer}.
While query-based methods \cite{wang2023exploring, lin2023sparse4d} have demonstrated exceptional performance in temporal modeling, our focus deviates from previous work in that we concentrate on extracting 3D voxel features instead of 2D BEV features. Consequently, we cannot rely on 2D queries to extract such 3D features.

\paragraph{\textbf{Height-Channel Transformation.}} Inspired by the Channel-to-Height module in FlashOCC \cite{yu2023flashocc}, we design a Height-Channel transformation module. We first perform a simple reshape operation along the channel and Z dimension, transforming the voxel feature $\textbf{V} \in R^{C\times Z\times H\times W} $ to BEV feature $\textbf{B} \in R^{C^{'}\times H\times W} $, where $C^{'} = C\times Z$. By employing this module, we temporarily compress the height dimension, thereby adapting it to query-based temporal modeling methods. The process can be shown as follows:
\begin{equation}
\left \{ \textbf{B}_{t-N}, ...,\textbf{B}_{t} \right \} = Height2Channel(\left \{ \textbf{V}_{t-N}, ...,\textbf{V}_{t} \right \}).
\end{equation}

\paragraph{\textbf{Temporal Transformer.}} Previous studies \cite{zong2023temporal, park2022time} have revealed the high complementary between short-term and long-term temporal fusion. Therefore, in our work, we model them independently to exploit their mutual complementarity and synergy. The temporal transformer is divided into two branches: short-term and long-term. Finally, a shallow MLP network is used to merge the features.

We define two grid-shape learnable BEV queries $\textbf{Q}^{short}_{t-m}$ and $ \textbf{Q}^{long}_{t-m} \in R^{C\times H\times W} $. The short-term query $\textbf{Q}^{short}_{t-m}$ is responsible for aggregating features from adjacent frames. For $\textbf{B}_{t-m}$, they are $\textbf{B}_{t-m-1}$ and $\textbf{B}_{t-m+1}$. We utilize deformable attention\cite{zhu2020deformable} to extract temporal information, which can enhance computational efficiency while providing a more fine-grained perception of input features. It is particularly suitable for capturing temporal information. Specifically, we generate a grid-shape set of reference points in BEV plane, denoted as $p$, in the t-m frame. Then, using the vehicle's motion information, we perform coordinate transformations on these reference points to obtain $p'$ in the coordinate system of the adjacent frames. The process can be represented as :
\begin{equation}
\hat{\textbf{B}}_{t-m}^{short} = DeformAttn(\textbf{Q}_{t-m}^{short},p',\left \{ \textbf{B}_{t-m-1}, \textbf{B}_{t-m+1} \right \} ).
\end{equation}
In this context, $DeformAttn(q,p,x)$ represents the deformable attention operator, where $q$ is the query, $p$ is the reference points and $x$ is the feature that interacts with the query.

The long-term branch is similar to the short-term branch, the difference lies in the fact that the long-term query $\textbf{Q}^{long}_{t-m}$ interacts with the entire BEV sequence, except $\textbf{B}_{t-m}$. It is worth noting that due to the long-term query interacting with a larger number of features, considering the trade-off between performance and efficiency, we perform dimensionality reduction on the features to make the calculations more efficient. The process can be represented as :
\begin{equation}
\hat{\textbf{B}}_{t-m}^{long} = DeformAttn(\textbf{Q}_{t-m}^{long},p',\left \{ \textbf{B}_{t-N},...,\textbf{B}_{t} \right \} - \textbf{B}_{t-m} ).
\end{equation}
After that, we concatenate the long-term and short-term BEV features together and pass them through a shallow MLP to output the final BEV feature.

\begin{equation}
\hat{\textbf{B}}_{t-m} = MLP(Concatenate(\hat{\textbf{B}}_{t-m}^{short},\hat{\textbf{B}}_{t-m}^{long} )).
\end{equation}

Finally, we apply the inverse Height-Channel transformation to revert the reconstructed BEV feature $\hat{\textbf{B}}_{t-m} \in R^{C' \times H\times W} $ back into the voxel feature $\hat{\textbf{V}}_{t-m} \in R^{C\times Z\times H\times W} $
\begin{equation}
\hat{\textbf{V}}_{t-m} = Channel2Height(\hat{\textbf{B}}_{t-m}).
\end{equation}

\subsection{Neural Rendering}
\label{subsec:Decoder}
To achieve unsupervised learning without any labels, we project the voxel features $\hat{\textbf{V}}_{t-m}$ back to 2D plane. Similar to previous work \cite{huang2023selfocc,yang2023unipad}, we employ differentiable volume rendering \cite{mildenhall2021nerf, sun2022direct} to synthesize color maps and depth maps, supervising them by original RGB images and depth maps constructed from LiDAR points. Inspired by recent advances in neural implicit surface reconstruction \cite{wang2021neus}, we represent a scene as an implicit signed distance field (SDF). The SDF encode the distance to the closest surface at every location as a scalar value. Compared to NeRF-like 3D representations, the use of SDF allows us to implicitly model 3D scenes based on the zero-level set, enabling our model to capture high-quality, fine-grained geometric information more effectively.

We generate multiple rays, and the $i$-th ray can be represented as $r_{i} (t)=o+td_{i} $, where $o$ and $d_{i}$ are the camera origin and direction of the $i$-th ray respectively. Then we sample $K$ points along the ray to obtain $ \left \{ p_{j} =o+t_{j}d_{i}| j=1,....,K \right \} $, where $p_{j}$ is the 3D coordinates, and $t_{j}$ is the corresponding depth along the ray. For discrete voxel features $\hat{\textbf{V}}_{t-m} $ and continuous 3D coordinate $p_{j}$, we employ bicubic interpolation (BI) to predict the corresponding feature $f_{j}$. Subsequently, we follow \cite{wang2021neus, oechsle2021unisurf} and employ a shallow MLP to get SDF value $s_{j}$ and geometry feature $g_{j}$. After that, we use another MLP network to obtain the corresponding colors for each sampled point. The entire process can be represented as :
\begin{equation}
f_{j} = BI(V, p_{j}), \quad   s_{j} = \Phi _{SDF} (f_{j}, p_{j}), \quad c_{j} = \Phi _{RGB} (f_{j}, p_{j}, d_{i}, n_{j}, g_{j}), 
\end{equation}
where $n_{j}$ is the gradient of the SDF value at the ray point $p_{j}$, $\Phi _{SDF}$ and $\Phi _{RGB}$ are two shallow MLP networks. Finally, to obtain the color and depth, we adopt the same approximation scheme as used in Neus \cite{wang2021neus}, this scheme samples $K$ points along the ray to compute the approximate pixel color and depth of the ray as:
\begin{equation}
\hat{C}_{i}= \sum_{j=1}^{K} T_{j}\alpha _{j}c_{j}, \quad \hat{D}_{i}= \sum_{j=1}^{K} T_{j}\alpha _{j}t_{j}, 
\end{equation}
where $T_{j}$ is the discrete accumulated transmittance defined by $T_{j} =  {\textstyle \sum_{k=1}^{j-1}}(1-\alpha _{k}) $, and discrete opacity value $\alpha_{j}$ which is the probability of the ray terminating between $p_{j}$ and $p_{j+1}$, can be defined by: 
\begin{equation}
\alpha_{j}=max(\frac{\sigma (s_{j}) - \sigma (s_{j+1})}{\sigma (s_{j})},0 ),
\end{equation}
where $\sigma (s_{j}) = (1+e^{-as_{j}})^{-1}$ is a parameterized sigmoid function with $a$ being a learnable parameter.

\subsection{Self Supervised Loss}
\label{subsec:Loss}
A straightforward way to supervise our model is to calculate pixel-wise RGB loss between rendered images and the training images. However, our experiments have shown that it does not work effectively. Depth supervision enhances the convergence of neural radiance fields, enabling faster and more effective learning of accurate geometry, especially with sparse views. Unlike the novel view synthesis task using dense supervision to improve the quality of rendered images, our task is to pre-train the backbone instead of synthesizing high-quality images. Therefore, we use sparse supervision to improve training efficiency. Following UniPAD \cite{yang2023unipad}, we employ a depth-aware sampling strategy to project the lidar point cloud onto multi-view images, selecting a pixel set with depths below a certain threshold, and we randomly select $M$ pixels from this set as supervision. This sampling strategy not only reduces computational complexity but also allows our model to focus on the foreground instead of irrelevant background elements, such as the sky. 

In summary, our loss function is defined as :
\begin{equation}
Loss = \frac{\lambda _{RGB}}{M} \sum_{i=1}^{M}\left | \hat{C}_{i} - C_{i}  \right |  +\frac{\lambda _{Depth}}{M} \sum_{i=1}^{M}\left | \hat{D}_{i} - D_{i}  \right | , 
\end{equation}
where $C_{i}$ and $D_{i}$ represent the pixel-wise ground-truth color and depth, respectively.

\section{Experiments}

\subsection{Dataset}
We evaluate the effectiveness of our work by utilizing the nuScenes \cite{caesar2020nuscenes} benchmark, which consists of  1000 scenes. The dataset is divided into training, validation, and testing splits, containing 700, 150, and 150 scenes, respectively. Each scene contains a 20-second video clip captured at a frame rate of 2 frames per second (FPS), along with 6 viewpoint images. In addition to 3D bounding box labels, the dataset provides High-Definition(HD) maps and information on vehicle motion states and camera parameters. To assess the 3D object detection performance, we follow the official guideline, using mean Average Precision (mAP) and nuScenes Detection Score (NDS). For BEV segmentation evaluation, we use IoU metric following CVT~\cite{zhou2022cross}. For vectorized HD map construction task, we use mAP metric following VectorMapNet~\cite{liu2023vectormapnet} and MapTR~\cite{liao2022maptr}.

\subsection{Implementation Details}
During the pre-training phase, we partially mask the inputs, focusing only on visible regions for feature extraction. To ensure effective mask, we first perform depth-aware sampling by selecting 512 rays per image view. Then, based on the selected pixel positions, we mask an image patch with a size of 16 centered around each selected pixel. After that, we mask the remaining visible regions using the configuration where masking size and ratio are configured to 32 and 0.3. The shape of the voxel feature $V$ is $128\times 128 \times 5$. For the voxel decoder, we generate the BEV query with resolution $128\times 128$, the dimension of the short-term query is 128, while the dimension of the long-term query is 64. We consider the current and 4 previous frames as a sequence and randomly select one frame from this sequence to serve as the reconstruction object. For the neural rendering network, following UniPAD \cite{yang2023unipad}, 512 rays per image view and 96 points per ray are randomly selected. The loss scale factors for $\lambda_{RGB}$ and $\lambda_{Depth}$ is 10. Unless otherwise specified, the image size is processed to $800 \times 450$ to facilitate training on RTX 3090 GPUs. The model undergoes training for 12 epochs using the AdamW optimizer with an initial learning rate of 2e-4 and 0.01 weight decay without the CBGS \cite{zhu2019class} and any data augmentation strategies.

\subsection{Comparison With Previous Pre-training Methods}
\paragraph{\textbf{Implementation.}}
In Table~\ref{tab:Comparison with other supervised or unsupervised pretraining methods}, we compare our method with previous pre-training methods to demonstrate the advantages of our work. Following UniPAD \cite{yang2023unipad}, we utilize ConvNeXt-S \cite{liu2022convnet} as the backbone for pre-training and conduct experiments on UVTR \cite{li2022unifying}. We use ConvNeXt-S pre-trained on ImageNet \cite{deng2009imagenet} as our baseline and examine the gains brought by various pre-training methods. DD3D \cite{park2021pseudo} utilizes depth estimation for pre-training, enhancing the model's perception of 3D geometric information. Spark \cite{tian2023designing} incorporates the MAE-style pre-training method into CNN networks, improving the model's overall grasp of the data. nuImages represents the image backbone is initialized with the pre-trained weights from MaskRCNN \cite{he2017mask} on the nuImages \cite{caesar2020nuscenes} dataset. In this method, 2D labels are utilized for supervision. FCOS3D \cite{wang2021fcos3d} is a widely used monocular 3D detector that employs 3D labels for supervision during the pre-training phase. UniPAD \cite{yang2023unipad} uses rendering-based pretext task to pre-train the model, similar to us. It is worth noting that in order to enable pre-training on the RTX 3090 GPUs, we lower the input image resolution to $800 \times 450$ during the training of UniPAD as well as our method, instead of using full-resolution images in the UniPAD paper. Additionally, pre-training is conducted for 12 epochs on the full dataset, and fine-tuning is conducted for 12 epochs on $50 \%$ of the dataset. 

\paragraph{\textbf{Results.}}
Upon leveraging our pre-trained model as the backbone, we observe a remarkable increase of 9.2 mAP and 6.6 NDS compared to the baseline. Furthermore, our model achieves an advancement over the previous state-of-the-art representation learning method UniPAD, with an improvement of 0.8 mAP and 1.1 NDS.
The experimental results indicate that even without using any labeled data for supervision, our method can surpass previous supervised or unsupervised approaches. This demonstrates the effectiveness of our approach, which synergizes temporal modeling with volume rendering. 

\begin{table}[htbp]
  \caption{Comparison with previous supervised or unsupervised pre-training methods. When adopt our pre-train model as the backbone, it brings 9.2 mAP and 6.6 NDS gains over the baseline. Moreover, \thename{} attains a 0.8 mAP and 1.1 NDS improvement over the previous state-of-the-art representation learning approach UniPAD~\cite{yang2023unipad}.
  }
  \label{tab:Comparison with other supervised or unsupervised pretraining methods}
  \centering
  \begin{tabular}{l|c|c|cc|c|c}
    \toprule
    \multirow{2}*{Model}  & \multirow{2}*{Backbone}  & \multirow{2}*{Pretrain model}  &\multicolumn{2}{c|}{Label}& \multirow{2}{*}{mAP}& \multirow{2}{*}{NDS}\\
    &&&2D &3D& \\

    \midrule
    
    UVTR & ConvNeXt-S & ImageNet & & & 23.0 & 25.2\\
    UVTR & ConvNeXt-S & DD3D & & & 25.1 & 26.9\\
    UVTR & ConvNeXt-S & SparK & & & 28.7 & 29.1\\
    UVTR & ConvNeXt-S & nuImages &\cmark & & 27.7 & 29.4\\
    UVTR & ConvNeXt-S & FCOS3D & & \cmark & 29.0 & 31.7\\
    UVTR & ConvNeXt-S & UniPAD & & & 31.1 & 31.0\\
    UVTR & ConvNeXt-S & \textbf{\thename{}} & & & \textbf{32.2} & \textbf{31.8}\\

  \bottomrule
  \end{tabular}
\end{table}

\subsection{Comparison With State-of-the-art Methods}
We extensively test our pre-trained model on various downstream tasks to ensure the generality of our approach.

\paragraph{\textbf{BEV Segmentation.}} In Table~\ref{tab:main result2}, we validate the effectiveness of our pre-train method on the BEV segmentation task. There are two commonly used evaluation settings for BEV segmentation. Setting 1 uses a 100m×50m area around the vehicle and samples a map at a 25cm resolution. This setting is adopted by \cite{roddick2020predicting} and widely employed as the benchmark against prior research. In contrast, the setting 2, popularized through \cite{philion2020lift}, used a 100m×100m area around the vehicle, with a 50cm sampling resolution.
ConvNetX-S~\cite{liu2022convnet} is employed as the backbone on the CVT~\cite{zhou2022cross}. Benefiting from our effective pre-training, it results in a significant $5.9\%$ IoU gain in Setting 1 and $8.7\%$ IoU gain in Setting 2.

\begin{table}[t!]
\caption{Comparisons of different methods with MIM4D pre-trained backbone on the nuScenes \textit{val} set in the BEV segmentation task. Setting 1 refers to the 100m×50m at 25cm resolution setting proposed by Roddick et al. \cite{roddick2020predicting}. Setting 2 refers to the 100m×100m at 50cm resolution setting proposed by Philion and Fidler \cite{philion2020lift}. Both settings evaluate the Intersection over Union (IoU) metric. }
  
  \label{tab:main result2}
  \centering
  \begin{tabular}{l|c|c}
    \toprule
    Model & Setting 1 (IoU\%) & Setting 2 (IoU\%)   \\
    \midrule

    PON \cite{roddick2020predicting} & 24.7 & - \\
    VPN \cite{pan2020cross}  & 25.5 & - \\
    STA \cite{saha2021enabling} & 36.0 & - \\
    Lift-Splat \cite{philion2020lift} & - & 32.1 \\
    FIERY Static \cite{hu2021fiery}  & 37.7 & 35.8 \\
    CVT \cite{zhou2022cross}  & 37.3 & 33.4 \\
    \textbf{CVT} \cite{zhou2022cross} \textbf{+ \thename{}}  & \textbf{39.5} & \textbf{36.3} \\

  \bottomrule
  \end{tabular}
\end{table}

\paragraph{\textbf{3D Object Detection.}} In Table~\ref{tab:main result}, we apply ResNet-50 \cite{he2016deep}, which is pre-trained by our method to the previous state-of-the-art methods and conduct experiments on the nuScenes validation set. We adopt PETR \cite{liu2022petr}, BEVDet4D \cite{huang2022bevdet4d}, and Sparse4Dv3 \cite{lin2023sparse4d} as baselines to demonstrate the effectiveness of our method. Benefiting from our effective pre-training, PETR outperforms the baseline over $ 2.6\% $ NDS, and BEVDet4D achieved a $3.5\%$ mAP gains over the baseline. Sparse4Dv3, which is currently the best method among vision-centric approaches, even though it incorporates fine-grained temporal modeling and employs depth estimation as an additional auxiliary task, our pre-training still yields a 0.6 NDS improvement on it.

\begin{table}[htbp]
  \caption{Comparisons of different methods  with MIM4D pre-trained backbone on the nuScenes \textit{val} set in the 3D object detection task. The results of Sparse4Dv3 are obtained from its official GitHub repository.
  }
  \label{tab:main result}
  \centering
  \begin{tabular}{l|c|c|c|c}
    \toprule
    Model & Backbone \quad & Image size \quad    & mAP \quad  & NDS \quad \\
    \midrule

    PETR \cite{liu2022petr}  & ResNet50 & 256 × 704  & 30.9  & 34.9\\
    \textbf{PETR + \textbf{\thename{}}}  & ResNet50 & 256 × 704   & \textbf{31.1}  & \textbf{35.8}\\
    BEVDet \cite{huang2021bevdet} & ResNet50 & 256 × 704 &    31.3 & 39.8\\
    BEVDet4D \cite{huang2022bevdet4d} & ResNet50 & 256 × 704 &  31.4  & 44.7\\
    \textbf{BEVDet4D + \textbf{\thename{}}} & ResNet50 & 256 × 704    & \textbf{32.5} & \textbf{45.0}\\
    BEVFormer v2 \cite{yang2023bevformer} & ResNet50 & 640 × 1600   & 42.0  & 51.8\\
    SOLOFusion \cite{park2022time}  & ResNet50 & 256 × 704   & 42.7  & 53.4\\
    VideoBEV \cite{han2023exploring}  & ResNet50 & 256 × 704  & 42.2  & 53.5\\
    StreamPETR \cite{wang2023exploring}  & ResNet50 & 256 × 704 & 43.2  & 53.7\\
    Sparse4Dv2 \cite{2305.14018}  & ResNet50 & 256 × 704  & 43.9  & 53.9\\
    Sparse4Dv3 \cite{lin2023sparse4d}  & ResNet50 & 256 × 704  & 46.3  & 56.4\\
    \textbf{Sparse4Dv3 + \textbf{\thename{}}}  & ResNet50 & 256 × 704  & \textbf{46.4}  & \textbf{57.0}\\
    
  \bottomrule
  \end{tabular}
\end{table}

\paragraph{\textbf{HD Map Construction.}} In Table~\ref{tab:main result3}, we evaluate the performance of our pre-trained model on the nuScenes \textit{val} set in the HD map construction task. High-definition (HD) map is high-precision maps specifically created for autonomous driving, consisting of instance-level vectorized representations of map features (pedestrian crossings, lane dividers, road boundaries, and so on). The HD map contains rich semantic information on road topology and traffic rules, which is essential for self-driving vehicle navigation.
MapTR \cite{liao2022maptr} is adopted as our baseline. With our well-pretrained backbone, MapTR, which uses BEVFormer  \cite{li2022bevformer} and GKT \cite{chen2022efficient} as map encoders, improved the mAP by $1.2\%$ and $1.4\%$ respectively, compared to the baseline.

\begin{table}[htbp]
  \caption{Comparisons of different methods with MIM4D pre-trained backbone on the nuScenes \textit{val} set in the HD map construction task. ``C'' and ``L'' respectively denote camera and LiDAR. $\dag$ indicates MapTR-tiny uses BEVFormer as BEV encoder, while $\star$ indicates MapTR-tiny uses GKT as BEV encoder.}
  
  \label{tab:main result3}
  \centering
  \begin{tabular}{l|c|c|c|c}
    \toprule
    Model & Modality & Backbone    & Epochs & mAP    \\
    \midrule

    HDMapNet \cite{li2022hdmapnet}   & C & EfficientNet-B0 \cite{tan2019efficientnet}   & 30  & 23.0\\
    HDMapNet \cite{li2022hdmapnet}   & L & PointPillars \cite{lang2019pointpillars}   & 30  & 24.1\\
    VectorMapNet \cite{liu2023vectormapnet}   & C & ResNet50   & 110  & 40.9\\
    VectorMapNet \cite{liu2023vectormapnet}  & L & PointPillars \cite{lang2019pointpillars}   & 110  & 34.0\\
    MapTR$^{\dag}$ \cite{liao2022maptr}  & C & ResNet50   & 24  & 48.7\\
    \textbf{MapTR$^{\dag}$} \cite{liao2022maptr} \textbf{+ \thename{}}    & C & ResNet50   & 24  & \textbf{49.3}\\
    MapTR$^{\star}$ \cite{liao2022maptr}   & C & ResNet50   & 24  & 50.0\\
    \textbf{MapTR$^{\star}$} \cite{liao2022maptr} \textbf{+ \thename{}}    & C & ResNet50   & 24  & \textbf{50.7}\\

  \bottomrule
  \end{tabular}
\end{table}

\subsection{Ablation Studies}
\label{subsec:Ablation Studies}
We conduct a series of ablation experiments to demonstrate the roles of different components in our model. In order to facilitate rapid iteration, in this section, we adopt UVTR \cite{li2022unifying} as the baseline with ResNet-50 \cite{he2016deep} as the backbone. Additionally, we downsample the input image resolution to $800 \times 450$, and conduct 12 training epochs on the full dataset in the pre-training stage and 12 training epochs on $50 \%$ of the dataset in the fine-tuning stage.

\paragraph{\textbf{The Length of The Time Window.}} In Table~\ref{tab:Ablation1}, we discuss the influence of the length of the input image frame sequence on our method. Specifically, when the length is 1, our method is like UniPAD. The experimental results demonstrate that as the length of the input image frame sequence increases, our method consistently exhibits improvements in mAP and NDS. As we fuse more frames into \thename{}, the mAP and NDS dramatically increase, improving by $8.8\%$ and $5.9\%$ from 1 to 5 frames. This supports our suspicion that the potential for localization is significantly increased by leveraging multiple frames over a longer time window, as opposed to using a single frame. It is worth noting that our method's performance does not saturate until the length of the time window reaches 5. This may indicate that, with GPUs having larger memory capacity and further increasing the length of the time window, there is potential to achieve better results with our method.

\begin{table}[htbp]
  \caption{Ablation study for the time window length. $N$ indicates the length of the time sequence. When the length of the time window increases from 1 to 5, the mAP and NDS dramatically increase, improving by $8.8\%$ and $5.9\%$.
  }
  \label{tab:Ablation1}
  \centering
  \begin{tabular}{l|c|c|c|c|c}
    \toprule
    Model & Backbone \quad & Image size \quad  & $N$ \quad   & mAP \quad  & NDS \quad \\
    \midrule
    UVTR  & ResNet50 & 450 $\times$ 800 & 1  & 18.2  & 22.2\\
    UVTR  & ResNet50 & 450 $\times$ 800 & 3  & 18.3  & 22.4\\
    UVTR  & ResNet50 & 450 $\times$ 800 & 4  & 18.8  & 22.8\\
    UVTR  & ResNet50 & 450 $\times$ 800 & 5  & \textbf{20.1}  & \textbf{23.5}\\

  \bottomrule
  \end{tabular}
\end{table}

\paragraph{\textbf{The Approach for Temporal Modeling.}} In Table~\ref{tab:Ablation2}, we investigate the impact of different temporal modeling methods on the performance of our model: 
\begin{itemize}
\item None. It represents that no temporal modeling is performed, and in this case, our method is like UniPAD. 
\item Warp-Cat. We perform coordinate transformation to align all the features in the sequence to the target frame. These transformed features are then concatenated and undergo temporal fusion using an MLP, similar to SOLOFusion \cite{park2022time}. 
\item Short-term. We only use short-term temporal transformer to extract temporal information.
\item Long-term. Similar to the Short-term method, only the long-term temporal transformer is used.
\item Long-term and Short-term. Both the long-term and short-term temporal transformer are utilized.
\end{itemize}
The results show that both the long-term and short-term temporal transformers bring performance improvements to the model when used separately, and we achieve the best performance when they are adopted together, demonstrating the complementarity between these two branches.

\begin{table}[htbp]
  \caption{Ablation study for temporal modeling. The best performance is achieved when long-term and short-term temporal transformers are adopted together, it demonstrates the complementary nature between these two branches.
  }
  \label{tab:Ablation2}
  \centering
  \begin{tabular}{l|c|c|c}
    \toprule
    Model & Method \quad   & mAP \quad  & NDS \quad \\
    \midrule
    UVTR  & None & 18.2  & 22.2\\
    UVTR  & Warp-Cat & 18.9  & 22.6\\
    UVTR  & Short-term & 18.3  & 22.4\\
    UVTR  & Long-term & 19.5  & 23.2\\
    UVTR  & Long-term and Short-term & \textbf{20.1}  & \textbf{23.5}\\
    
  \bottomrule
  \end{tabular}
\end{table}

\section{Conclusion}
In this paper, we introduce a new representation learning method \thename{} for autonomous driving.
\thename{} is a dual masked image modeling (MIM) framework in both the temporal and spatial domains. Given masked multi-view videos as input, it leverages continuous scene flow to construct the dropped voxel feature, and then provide pixel-wise supervision through 3D differential volume rendering.
Our empirical results show that MIM4D outperforms previous pre-training methods and exhibits superior performance across various downstream tasks. We believe this novel method for representation learning could provide a new perspective on scalable autonomous driving. 

\bibliographystyle{splncs04}
\bibliography{main}
\end{document}